\documentclass[conference]{IEEEtran}
\IEEEoverridecommandlockouts
\usepackage{cite}
\usepackage{amsmath,amssymb,amsfonts}
\usepackage{algorithmic}
\usepackage{graphicx}
\usepackage{textcomp}
\usepackage{xcolor}
\usepackage{multirow}
\usepackage{float}
\usepackage{stfloats}
\usepackage{changes}
\begin{document}
{
\title{Multi-level Similarity Learning for Low-Shot Recognition
{\footnotesize }
\thanks{This work was supported in part by the National Natural Science Foundation of China (No. 41576011, U1706218).
}
}
\author{Hongwei Xv, Xin Sun, Junyu Dong, Shu Zhang, Qiong Li\\
	Department of Computer Science and Technology\\
	Ocean University of China\\
	Qingdao, China\\ 
    \{xhw, liqiong\}@stu.ouc.edu.cn, \{sunxin, dongjunyu, zhangshu\}@ouc.edu.cn\\
}
\maketitle

\begin{abstract}
Low-shot learning indicates the ability to recognize unseen objects based on very limited labeled training samples, which simulates human visual intelligence. According to this concept, we propose a multi-level similarity model (MLSM) to capture the deep encoded distance metric between the support and query samples. Our approach is achieved based on the fact that the image similarity learning can be decomposed into image-level, global-level, and object-level. Once the similarity function is established, MLSM will be able to classify images for unseen classes by computing the similarity scores between a limited number of labeled samples and the target images. Furthermore, we conduct 5-way experiments with both 1-shot and 5-shot setting on Caltech-UCSD datasets. It is demonstrated that the proposed model can achieve promising results compared with the existing methods in practical applications.
\end{abstract}


\section{Introduction}
Deep convolutional neural networks (ConvNets) have achieved great success on visual recognition tasks, such as face recognition \cite{Calefati2018}, object detection \cite{Lin2017} and many others. Among those tasks, the success of visual recognition systems greatly relies on the large quantities of annotated data, e.g., ImageNet \cite{JiaDeng2009}. It usually needs thousands of labeled examples for each class to saturate ConvNets' performance \cite{Wang2018}. However, in practice, it might be extremely expensive or infeasible to obtain sufficient labeled images. On the other hand, the human perception system can easily understand unseen concepts with little knowledge training, especially for the domain-specific low-shot tasks (e.g., Ornithologists will recognize new birds more quickly than ordinary people). This challenge of learning new concepts from a very limited number of labeled samples often is referred to as \textit{low-shot learning} or \textit{few-shot learning}. And this is what this paper focuses on.
\begin{figure}[t]
	\centerline{\includegraphics[width=1.0\linewidth]{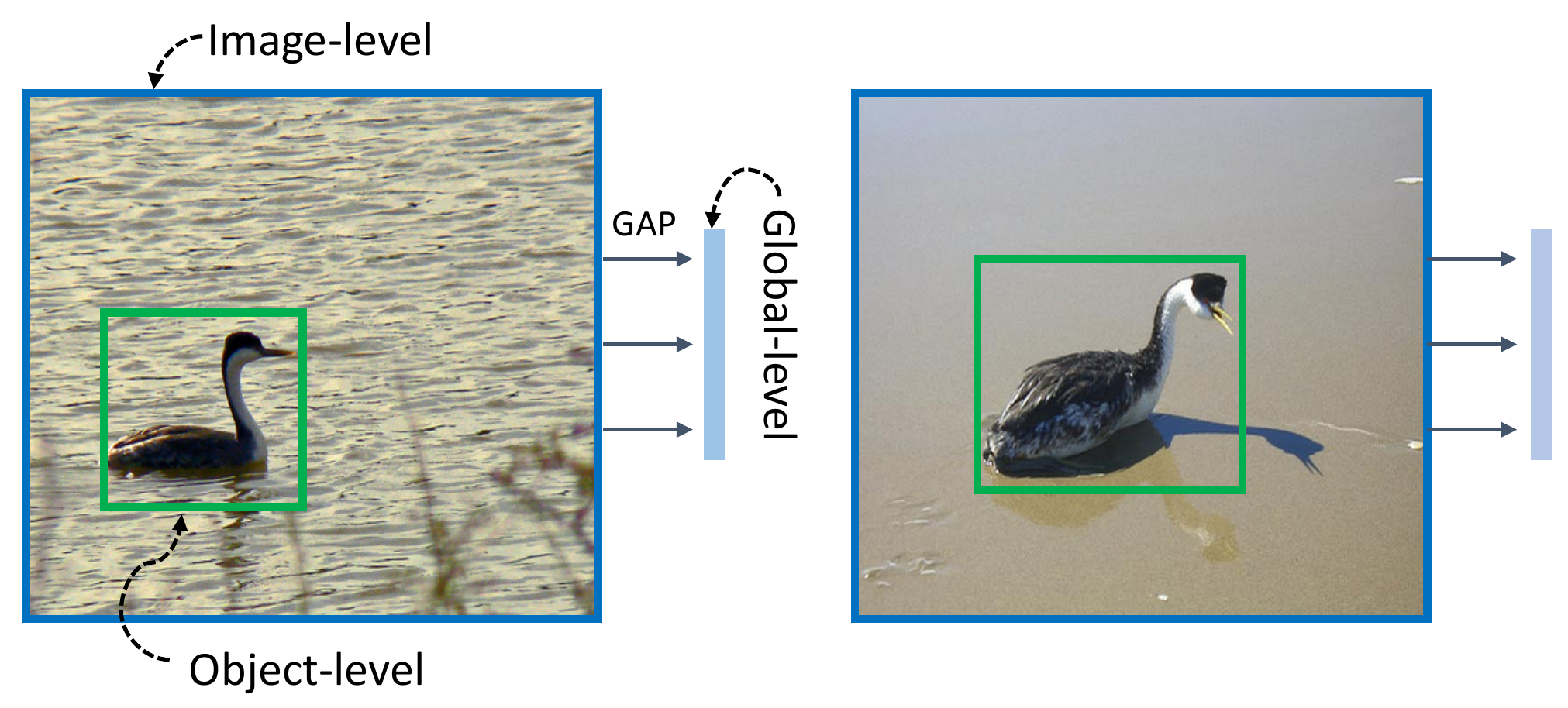}}
	\caption{Three different levels of similarity between images.}
	\label{threelevel}
\end{figure}

One way to address this problem is to utilize the strategy of transfer learning---a fine-tuning network using a few labeled samples for a new classification category. However, the whole network can be broken down by very few data due to the overfitting. With an urgent need on low-shot learning in practice, there are two research approaches fall under the umbrella of {\it meta-learning} \cite{Finn2017, Ravi2017} and {\it metric learning} \cite{Vinyals2016,Brown}. Approaches of meta-learning often train a meta-learner, which learns to directly map between the training sets and the testing examples for classification. Meanwhile, there are other meta-learning methods such as Meta-SGD \cite{Li2017} generate an adaptive learning rate by adjusting differentiable learner in one or few steps for training the classifier. Metric-learning tries to solve the low-shot problems by learning an embedding function to map a few numbers of labeled images into an embedding space. Then the learned space can classify test images via a nearest neighbor algorithm based on the distance measurements such as Euclidean or cosine distance.

In this paper, we adopt the metric learning method. The proposed strategy is based on the following facts. When a person observes the correlation between two images, they first directly get the concept of the overall similarity between the two images (image-level information as shown in Fig. \ref{threelevel}). Then his/her attention will move to the target object existed in the images (object-level information as shown in Fig. \ref{threelevel}) to measure the target object similarity. In addition, the global-level information is a compressed version of overall image-information, it is a complement to image-level information. Inspiringly, we propose a Multi-level Similarity Model (MLSM) that performs a low-shot recognition by learning how to compare the test images against low-shot labeled samples. The element-sum operation will be performed on these three levels of similarities. The similar function is achieved by a fully connected network.

The main contribution of this paper is to propose a multi-level relation model obtained from training classes. It computes relation scores between the test images and the limited number of examples. The experiment shows that the proposed model could achieve promising result compare with the state-of-the-art methods on Caltech-UCSD Birds datasets.  
\begin{figure*}[t]
	\centerline{\includegraphics[width=1.0\linewidth]{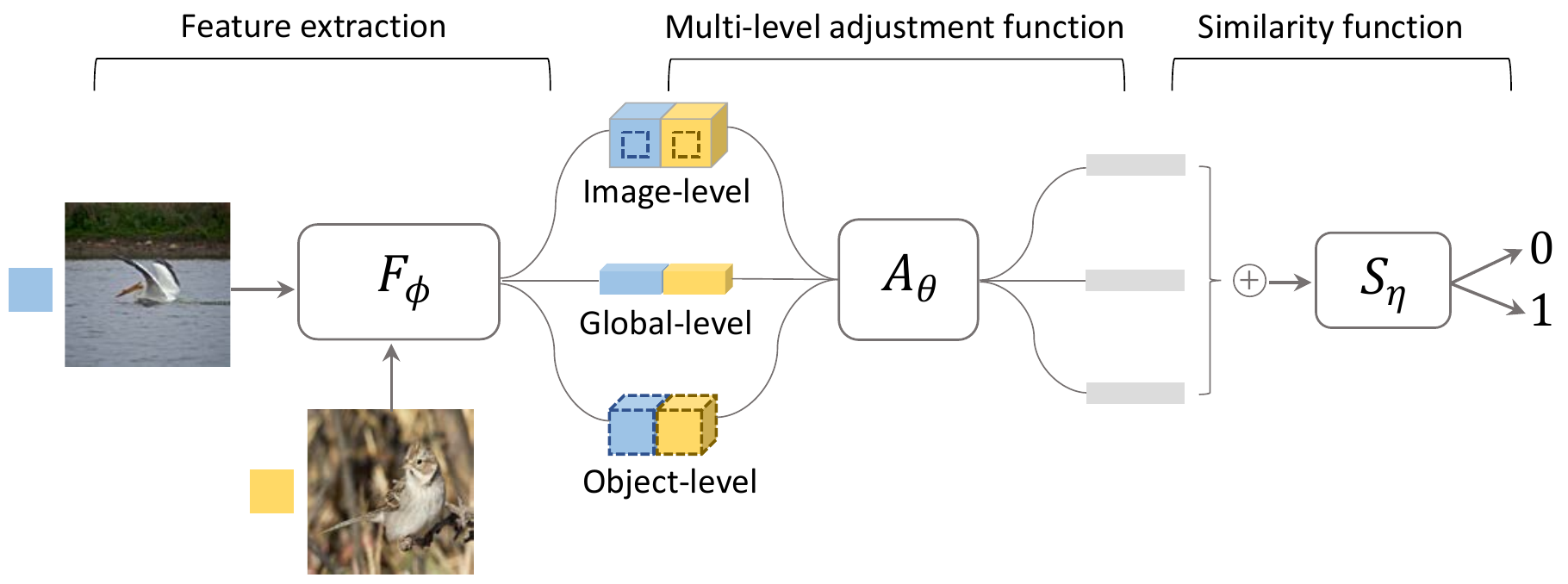}}
	\caption{The overall architecture of MLSM. The blue and yellow small squares represent different images, respectively, they can be the same category or not. }
	\label{allstruc}
\end{figure*}
\begin{figure*}[t]
	\centerline{\includegraphics[width=1.0\linewidth]{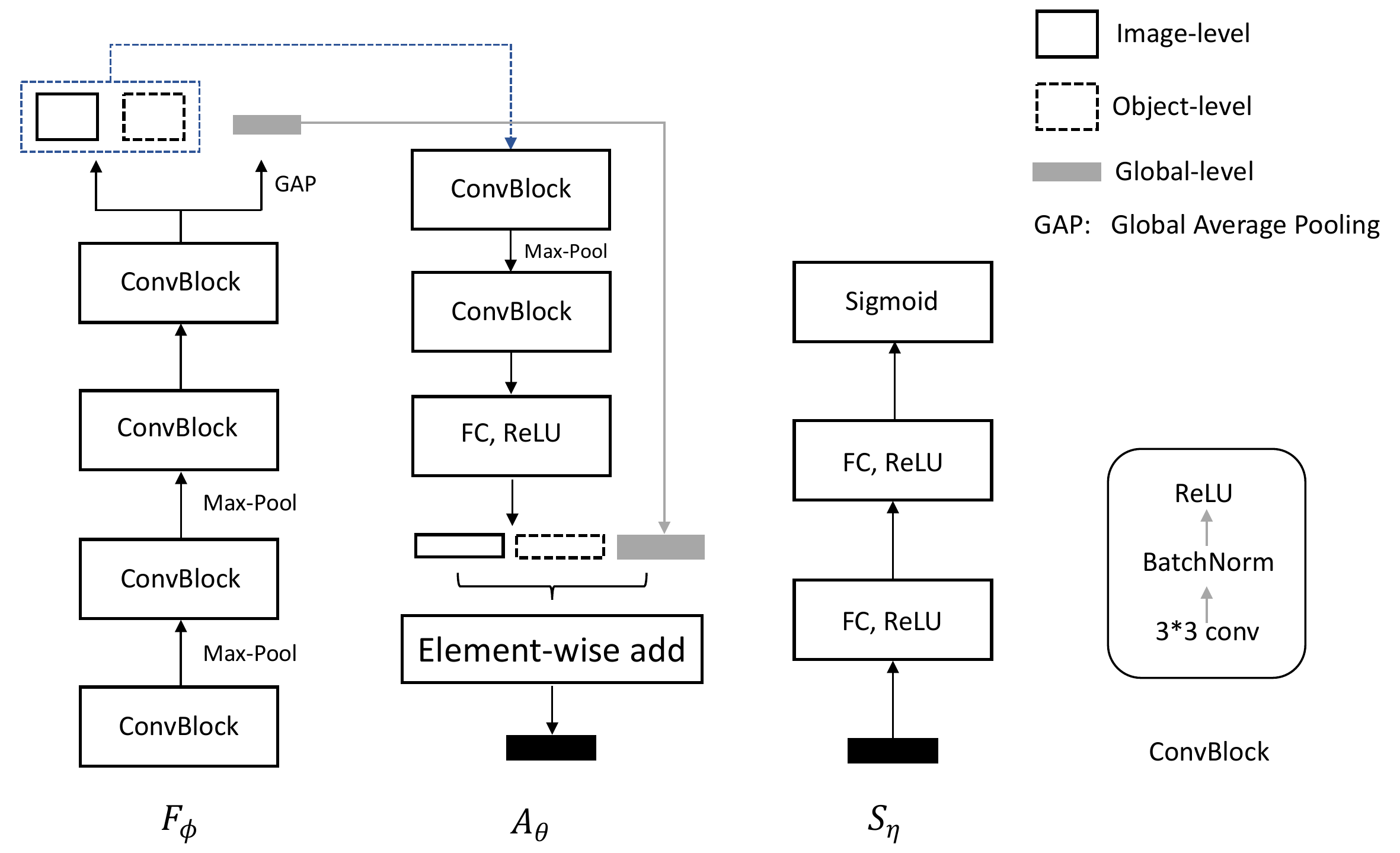}}
	\caption{Details of the architecture of MLSM.}
	\label{extract}
\end{figure*}

\section{Related Work}
\subsection{Few-shot Learning}
Amongst representation learning approaches, metric-learning is widely adopted, such as siamese networks \cite{Koch2015} or triplet loss \cite{Schroff2015}. They are used to minimize the distance between the inputs and the samples with the same label while keeping the distances of samples with different labels far apart. Such loss function has shown benefits in face identification and low-shot learning \cite{Koch2015, Zhang2015}. More generally, metric-learning based approaches, such as Matching Networks employs a differentiable nearest neighbor classifier over the learned representations of the training examples in order to classify an unlabeled example. Prototypical Networks \cite{Snell2017} handles the problems by calculating the Euclidean-distance between the embedding points of test samples and prototype representation of labeled samples. Relation Networks \cite{Sung2017} uses concatenated feature maps from the query and labeled images to distinguish the similarity and dissimilarity. Those methods including the proposed one follow the training strategy of meta-learning, which is to sample small training sets and query sets from base-train classes and feed the sampled training set to the learner for a classifier. It then computes the loss of the classifier on the sampled query set.
\subsection{Visualizing CNNs}
In addition, the object-level information is essentially the focused-regions in the image. The focused-region is usually captured by some methods like visualizing ConvNets(CNN). A number of previous works \cite{Zhou, Selvaraju2017} have visualized CNN predictions by highlighting 'important' pixels (i.e., usually some areas with discriminative features, such as the area containing the object). Class Activation Mapping (CAM) \cite{Zhou} proposed to localize object by modifying CNN architectures. It replaced the fully-connected layers with convolutional layers and global average pooling \cite{Lin} to obtain the focused-regions. Gradient-weighted Class Activation Mapping (Grad-CAM) \cite{Selvaraju2017} was applicable to a significantly broader range of CNNs without modifying models anymore. Grad-CAM utilized the gradients of any target concept (for example the logits for a caption) and fed it into the final convolutional layer to produce a coarse localization map, which highlighted the important regions in the image for prediction tasks. The object-level module in our work is inspired by the localization concept from Grad-CAM.

\subsection{Similarity Learning}
Amongst similarity learning, the ability to compare between images is one of the most fundamental operations among all of computing. Classic per-pixel measures, such as $L_2$ Euclidean-distance are insufficient for assessing structured images. How to measure two similar images in a way that coincides with human judgment is a longstanding goal \cite{Wang2004}. Richard Zhang et al. \cite{Zhang2018} systematically evaluated the deep features that are obtained by averaging the information across spatial dimension and across all layers. For the proposed model, different levels of deep features are utilized to measure the similarities rather than averaging all layers' outputs.

\section{MULTI-LEVEL SIMILARITY MODEL}\label{method}
\subsection{Problem Definition}
Assuming that there is a base train dataset $\mathcal{D}_{base}$, a novel test dataset $\mathcal{D}_{novel}$, where $\mathcal{D}_{base} \bigcap \mathcal{D}_{novel} = \emptyset$. Following the meta-learning normal training strategy, in each training iteration, we randomly sample $C$ classes with $K$ labeled samples (called {\it {$C$-way $K$-shot}}) from $\mathcal{D}_{base}$. These samples serve as support set $\mathcal{S}=\{(x_i,y_i)\}_{i=1}^{K*C} $. Then a fraction of the remainder in those $C$ classes is sampled to serve as a query set $\mathcal{Q} = \{(x_i,y_i) \}_{j=1}^N$. The loss of the classifier is computed on the $\mathcal{Q}$ to optimize the proposed model by back-propagation. During the testing of the proposed model, the $\mathcal{S}$ and $\mathcal{Q}$ are sampled from $\mathcal{D}_{novel}$ similarly to the above training procedure. The $\mathcal{Q}$ of $\mathcal{D}_{novel}$ can be classified without further updating the model.

As shown in Fig. \ref{allstruc}, the proposed model consists of three parts: a feature extraction module $F_{\phi}$, an adjustment function module $A_\theta$ and a similarity function module $S_\eta$. Given two images $\bf{a} \in \mathcal{S}$ and $\bf{b} \in \mathcal{Q}$ (or $\bf{b} \in \mathcal{S}$), the feature extractor $F_{\phi}$ will produce a pair of image-level feature maps {${\mathcal{I}}_{\bf{a}}$} and ${\mathcal{I}}_{\bf{b}}$. Object-level of $\bf{a}$ and $\bf{b}$ are also fed into the feature extractor $F_{\phi}$. Then ${\mathcal{I}}_{\bf{a}}$ and ${\mathcal{I}}_{\bf{b}}$ will generate a pair of global-level feature vectors ${\mathcal{G}}_{\bf{a}}$ and ${\mathcal{G}}_{\bf{b}}$ through the global average pooling (GAP). After processed by the adjustment function module $A_\theta$, three-level feature vectors will be obtained through the unified channels with the element-sum operation to integrate the multi-level feature representations. Then $\mathcal{C}$( (${\mathcal{I}}_{\bf{a}}$ $\oplus$ ${\mathcal{O}}_{\bf{a}}$  $\oplus$ ${\mathcal{G}}_{\bf{a}}$), (${\mathcal{I}}_{\bf{b}}$ $\oplus$ ${\mathcal{O}}_{\bf{b}}$  $\oplus$ ${\mathcal{G}}_{\bf{b}}$)) will be sent to $S_\eta$ to learn similar scores (1 represents the same class while 0 represents the different), where $\mathcal{C}(\cdot,\cdot)$ means the concatenation of feature vectors in depth. For $K$-shot, where $K > 1$, the feature vectors of $K$ labeled samples are averaged to serve as the representation of this class. The detailed structure of the three parts is shown in Fig. \ref{extract}.

\begin{figure}[t]
	\centerline{\includegraphics[width=1.0\linewidth]{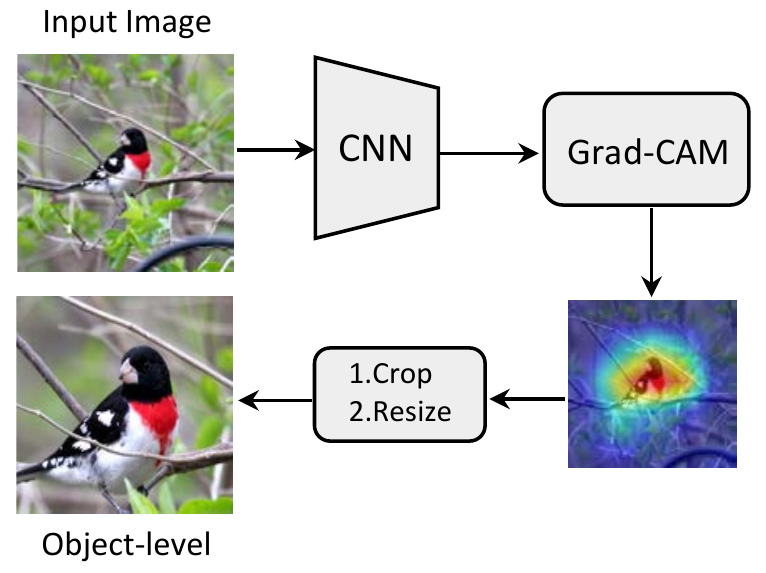}}
	\caption{A flowchart for generating object-level areas.}
	\label{object_level}
\end{figure}
\begin{figure}[t]
	\centerline{\includegraphics[width=1.0\linewidth]{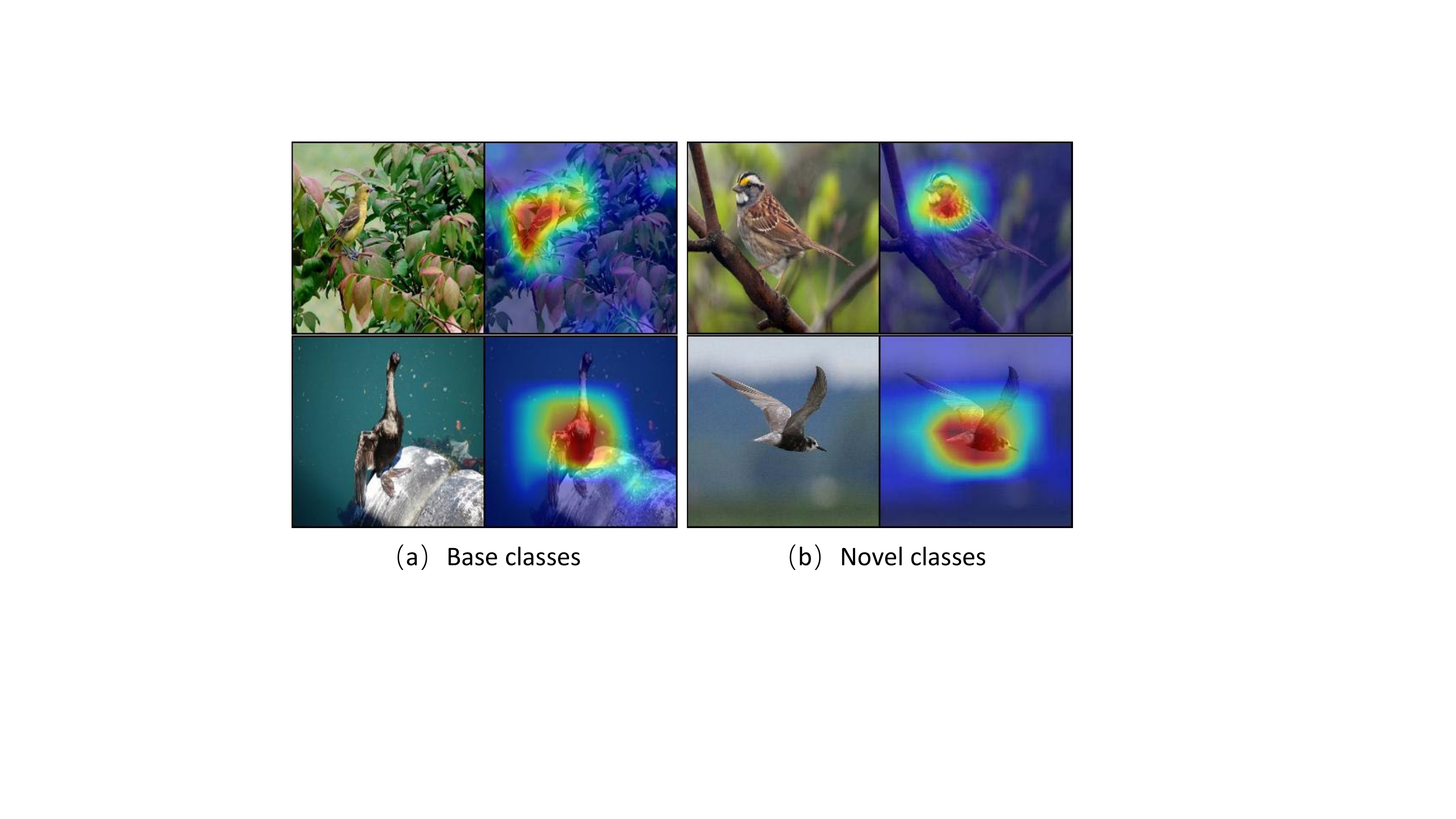}}
	\caption{The object-level areas generated by Grad-CAM. We can observe that the novel classes in (b) also can locate the coarse focus regions even it has never seen.}
	\label{cam}
\end{figure}
\subsection{Image-level Similarity}
Image-level information refers to the feature map of the whole image obtained by feature extractor $F_{\phi}$. For the proposed model, $F_{\phi}$ consists of four convolutional blocks, as shown in Fig. \ref{extract}. More specifically, each convolutional block contains a $3 \times 3$ convolution with 64 filters, batch normalization and a non-linear activation function (ReLU). Both training and testing samples are uniformly resized into $84 \times 84$. Feature maps at this level contain the highest-order semantic information of the image. Therefore, image-level is the best choice for similarity learning between images. A simple rule is to concatenate feature maps from $\bf{a}$ and $\bf{b}$ pair-wisely, i.e.
\begin{equation} \label{imageC}
\mathcal{C}({\mathcal{I}}_{\bf{a}}, {\mathcal{I}}_{\bf{b}}) = \{Concat( {\mathcal{I}}^i_{\bf{a}}, {\mathcal{I}}^i_{\bf{b}} ) \}_{i=1}^{M*N}
\end{equation}
where $M*N$ represents the size of the last feature map. Once two feature maps are concatenated together, the channel information contained in each pixel of the last feature map will be compared, like Relation Networks did \cite{Sung2017}.

\subsection{Object-level Similarity}

The final goal of low-shot learning is to classify different types of objects. The existing studies have shown that learning from object-areas could help for recognition tasks at image-level \cite{Fu2017, Li}. Inspiredly, we assume that there may exist some discriminative regions in the images that are beneficial to low-shot tasks. The discriminative region is the focused-area in the image. It usually contains all the information of the objects rather than the background. We call the focus-area as object-level area. To acquire the object-level area, Li et al. \cite{Li} proposed a zoom network which utilized the candidate region to crop the original images. Wei et al. \cite{Wei2019} adopted the unsupervised object discovery and co-localization mechanism by deep descriptor transformation to discover the object-level area. Differently, to be more efficient, we utilize Grad-CAM to get the object-level area. A coarse localization map highlighting the important regions in the image for predicting the concept could be calculated by:

\begin{equation}\label{grad}
L_{Grad-CAM}^c=ReLU(\sum\limits_k \alpha_k^c A^k)
\end{equation}
where $\alpha_k^c= \frac{1}{Z} \sum\limits_{i}\sum\limits_{j}\frac{\partial y^c}{\partial A_{ij}^k}$ denotes the weight of the $k$-th feature map for category $c$ and $Z$ is the number of feature map's pixels. $A^k_{ij}$ represents the pixel value at the location of $(i,j)$ of the $k$-th feature map, and $y^c$ is the classification score corresponding to the $c$-th class.

For this step, as shown in Fig. \ref{object_level}, we first train a ConvNets with the existing base data $\mathcal{D}_{base}$. The ConvNets is only used to generate object-level areas rather than extracting any features. Then we use the Grad-CAM to generate a  coarse object-level area. We further scale and zoom the target areas into a size of $84 \times 84$, which is the same as the original images. Nevertheless, there are still some problems when applying the proposed method, i.e. we do not know the label (corresponds to $c$ in equation (\ref{grad}) ) of new categories to generate the object-level areas. In this case, we utilize Grad-CAM directly and look for the closest class to $\mathcal{D}_{base}$ to locate the coarse object-level area, as shown in Fig. \ref{cam}.

\subsection{Global-level Similarity}
For the image-level feature, each of the learned filters operates with a local receptive field. It is thus unable to exploit contextual information outside of this region. In order to capture the global-level similarity, global spatial information is applied to the channel descriptor. This is achieved by the feature extractor $F_{\phi}$, which employs global average pooling after the last convolutional layer. The output features of $\bf{a}$ (or $\bf{b}$) are as follows:

\begin{equation}\label{gap}
{\mathcal{G}}_{\bf{a}} ( {\mathcal{G}}_{\bf{b}} ) = \frac{1}{H*W} \sum_{x,y} \alpha_{(x,y)}.
\end{equation}
where $\alpha_(x,y)$ denotes the $C$-dimensional slice of $\alpha$ at spatial location $(x,y)$. 
The global-level similarity of images $\bf{a}$ and $\bf{b}$ can be directly calculated by:

\begin{equation}\label{gl}
\mathcal{C}({\mathcal{G}}_{\bf{a}}, {\mathcal{G}}_{\bf{b}}) = \{Concat( {\mathcal{G}}^i_{\bf{a}}, {\mathcal{G}}^i_{\bf{b}} ) \}_{i=1}^{K}
\end{equation}
where $K$ denotes $K$-dimensional feature vector after GAP. In addition, due to the three different levels, the feature channel sizes are not unified. As shown in Fig. \ref{extract}, we then use an adjustment function $A_{\theta}$ to align the feature vectors of three levels for the element-sum operation. $A_{\theta}$ consists of two convolutional blocks and a fully-connected layer.

Overall, the multi-level similarity is calculated as follows:
\begin{equation}\label{ml} Similarity =  S_\eta \big( Concat( ({\mathcal{I}}_{\bf{a}} \oplus {\mathcal{O}}_{\bf{a}}  \oplus {\mathcal{G}}_{\bf{a}}), ({\mathcal{I}}_{\bf{b}} \oplus {\mathcal{O}}_{\bf{b}}  \oplus {\mathcal{G}}_{\bf{b} })) \big)
\end{equation}
where $S_{\eta}$ denotes the similarity function. It uses Sigmoid activation function to force the network to learn the concept of similarity and dissimilarity.
\begin{table*}[t]\large
	\centering
	\caption{Average test set classification accuracy on Caltech-UCSD Birds.}
	\begin{tabular}{|c|c|c|c|c|}
		\hline		
		\multirow{2}{*}{\textbf {Methods}} &\multicolumn{4}{|c|}{ \textbf{Caltech-UCSD Birds ( \% )}} \\
		\cline{2-5} 
		& \textbf{\textit{Fine Tune}}& \textbf{\textit{5-way 1-shot}}& \textbf{\textit{5-way 5-shot}} & \textbf{\textit{Distance}}\\
		\hline
		{	MAML \cite{Finn2017}}  &{Y}& 38.43& {59.15} & {-}\\
		\hline
		META-LEARN LSTM \cite{Li2017}&N&40.43&49.65&- \\
		\hline
		Matching Nets \cite{Vinyals2016}&N&49.34&59.31&Cosine \\
		\hline
		PROTO-Nets \cite{Snell2017}&N&45.27&56.35&Euclid.\\
		\hline
		RELATION-Nets \cite{Sung2017}&N&46.69&{55.86}&Image-level deep metric\\
		\hline
		MACO \cite{Hilliard2018}&N&60.76&\bf{74.96}&-\\
		\hline
		MLSM&N&\bf{64.50}&70.50&Multi-level deep metric \\
		\hline
	\end{tabular}
	\label{tab1}
\end{table*}

\begin{figure}[t]
	\centerline{\includegraphics[width=1.0\linewidth]{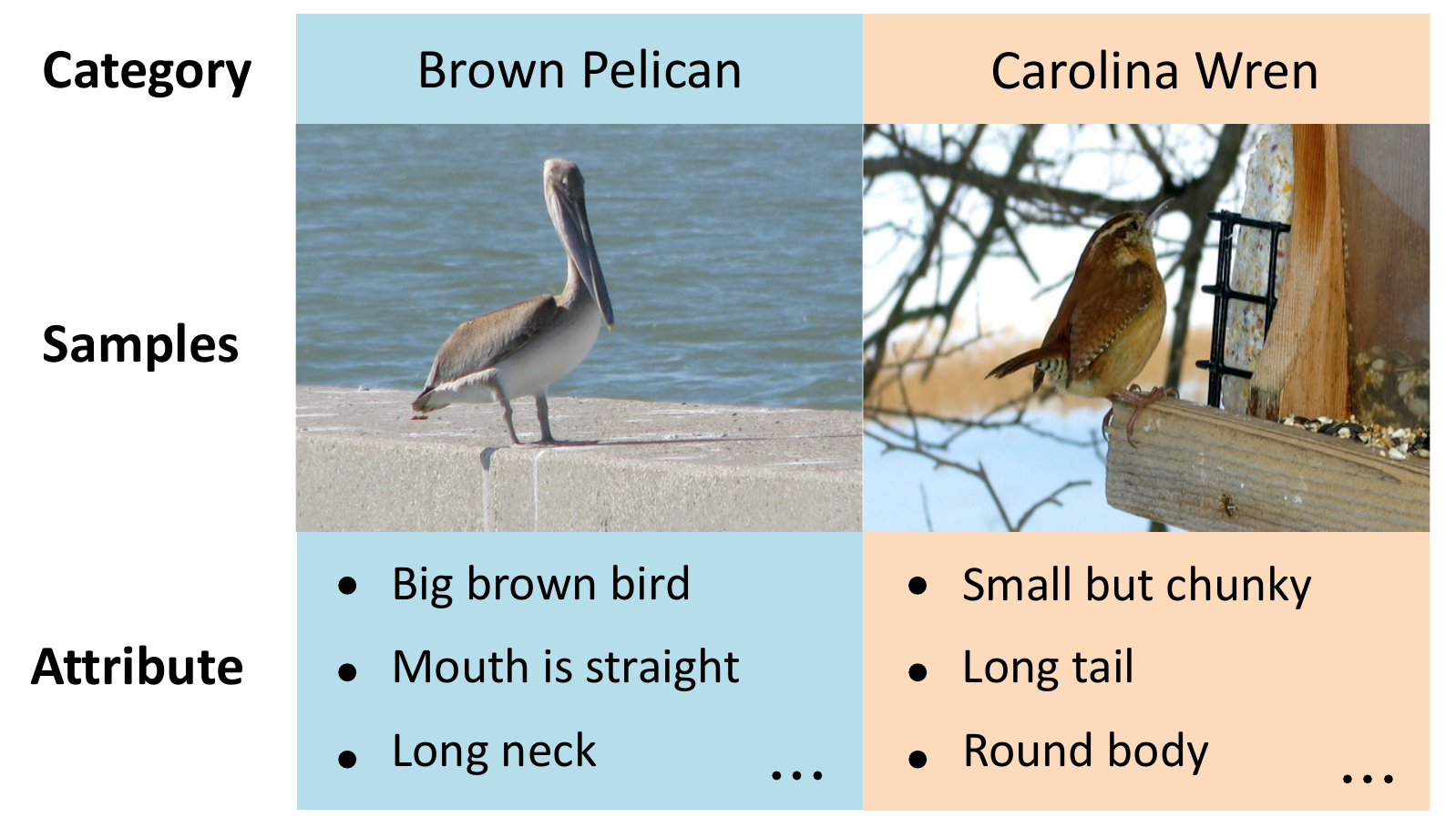}}
	\caption{Two categories randomly sampled from the Caltech-UCSD Birds dataset.}
	\label{cub}
\end{figure}
\begin{table}[t]
	\caption{Performance comparison on different levels.}
	\begin{center}
		\begin{tabular}{|c|c|c|}
			
			\hline
			\bf{ \quad Different level  \quad} &\bf{\quad1-shot (\%) \quad} & \bf{ \quad5-shot (\%) \quad} \\
			\hline
			\bf{ \quad $\mathcal{I} $  \quad} &{\quad 50.0 \quad} & { \quad59.0 \quad} \\
			\hline
			\bf{ \quad $\mathcal{I}+\mathcal{G}$ \quad} &{\quad 51.0 \quad} & { \quad61.0 \quad} \\
			\hline
			\bf{ \quad $\mathcal{I}+\mathcal{G}+\mathcal{O} $\quad} &\bf{\quad 64.5 \quad} & \bf{ \quad 70.5 \quad} \\
			\hline
			
		\end{tabular}
		\label{tab2}
	\end{center}
\end{table}

\section{Experiments}
\subsection{Datasets and Experimental Procedure}

{\bf{Datasets: }} we evaluate the proposed MLSM on the dataset of Caltech-UCSD 200-2011 \cite{Wah2011}. It consists of 11788 bird images in 200 categories. The samples of this dataset are shown in Fig. \ref{cub}. Each category has some semantic descriptions. We further divide the dataset into three parts: 100 classes for training (i.e. $\mathcal{D}_{base}$), 50 for validation, and 50 for testing(i.e. $\mathcal{D}_{novel}$). For the proposed model, the input images (including object-level areas) are resized to $84 \times 84$ pixels without any data augmentations (e.g. randomized rotation, and horizontal flipping). We use Adam solver \cite{Kingma2015} with an initial learning rate of 0.001, and annealed by half for every 100,000 episodes.

\subsection{Experiments on CUB-200-2011}
As shown in Table \ref{tab1}, it can be seen that the proposed model achieves promising results compared with other methods. The accuracy is computed by averaging over 100 randomly generated episodes from the testing set. Each episode contains 5 classes with 200 query images. Our model could achieve state-of-the-art performance on the 5-way 1-shot setting. At the same time, there are also some gaps between our method and MACO \cite{Hilliard2018} on 5-way 5-shot setting. Compared with the image-level deep metric method RelationNets \cite{Sun2017}, it can be noticed that the multi-level metric could be beneficial to learn a better similarity function with no additional training set.  We also compared the impact of different levels metric on the accuracy, from table \ref{tab2}, we can see that object-level representations bring great improvement both on 1-shot and 5-shot setting.

\section{Conclusion}
This paper introduces a novel multi-level similarity learning model (MLSM) for low-shot tasks. MLSM learns a similarity function from image-level, global-level, and object-level which coincides with human judgment to some extent.  The performance is evaluated on Caltech-UCSD 200-2011 dataset by comparing query and support samples. It achieves promising results compared with the existing methods. We believe that our method can act as a valuable complement to low-shot learning.
\section*{Acknowledgment}
This work was supported in part by the National Natural Science Foundation of China (No. 41741007, 41576011, U1706218)，the Key Research and Development Program of Shandong Province (no.GG201703140154) and Applied Basic Research Programs of Qingdao (no. 18-2-2-38-jch).
\cite{Xiao2017}
\bibliographystyle{IEEEtran}
\bibliography{xu}

\end{document}